\documentclass[letterpaper]{article}
\usepackage{iccc}
\usepackage[T1]{fontenc}
\usepackage{epigraph}
\usepackage{graphicx}
\usepackage{subfig}  
\usepackage{booktabs}
\usepackage{makecell}
\usepackage{hyperref}
\usepackage{times}
\usepackage{helvet}
\usepackage{courier}

\title{Reading Between the Lines: Can LLMs Discover the Question Behind the Text?}

\author{
Claudiu Creanga$^{2, 3}$, Liviu P. Dinu$^{1, 3}$\\ 
  $^1$ Faculty of Mathematics and Computer Science, \\
  $^2$ Interdisciplinary School of Doctoral Studies, \\
  $^3$ HLT Research Center, \\
  University of Bucharest, Romania\\
  \small
{\tt claudiu.creanga@fmi.unibuc.ro, ldinu@fmi.unibuc.ro}  \\
}

\begin{document} 
\maketitle
\begin{abstract}
This paper introduces ``question archaeology'', a specific evaluation task focused on inferring the single, authentic "genesis question" that motivated the creation of a complete text. Distinct from question generation, which targets any plausible question, or discourse frameworks that model utterance-level acts, our task assesses a model's grasp of authorial intent. We present a new dataset of commissioned texts paired with their original research questions and plausible distractors. Our evaluation of both proprietary models, like Gemini Flash and Pro, as well as open source models like Mistral and Qwen, reveals significant progress in this task, with the newer versions outperforming the earlier ones, while BERT-based models performed poorly. Notably, our findings indicate that current LLMs surpass human performance on this task, suggesting advanced understanding of authorial intent. This capability has important implications for AI's role in tasks requiring nuanced interpretation of human communication. Our work thus provides a new framework and a challenging benchmark for future models.

\end{abstract}

\epigraph{Judge a man [model] by his questions rather than by his answers.}{Voltaire}

\section{Introduction}

Every meaningful text that was written begins as an answer to a question - this is the heart of Bakhtin's dialogic theory \cite{Bakhtin1981}. This dialogic nature of text suggests that written discourse is fundamentally responsive, emerging as a reply to preceding utterances and anticipating future responses within an ongoing conversation. Understanding the questions that prompt textual creation is therefore fundamental to fully comprehend the meaning, purpose, and context of any written work. Despite this intrinsic relationship between questions and texts, models have traditionally been evaluated on their ability to answer questions about texts, rather than their capacity to identify the questions that motivated those texts in the first place. Traditional benchmarks in natural language understanding focus primarily on question answering, summarization, and classification tasks — all of which assess a model's ability to process and extract information from given content. However, these evaluations fail to address a more fundamental aspect of textual understanding: the ability to recognize \textbf{why} a text was created. This goes beyond simple topic classification, which identifies \textbf{what} a text is about. Our task aims to infer the why—the specific authorial intent or underlying question that prompted the text's creation, which is a more nuanced challenge.

\begin{figure}[htbp]
    \centering
    \includegraphics[width=1\linewidth]{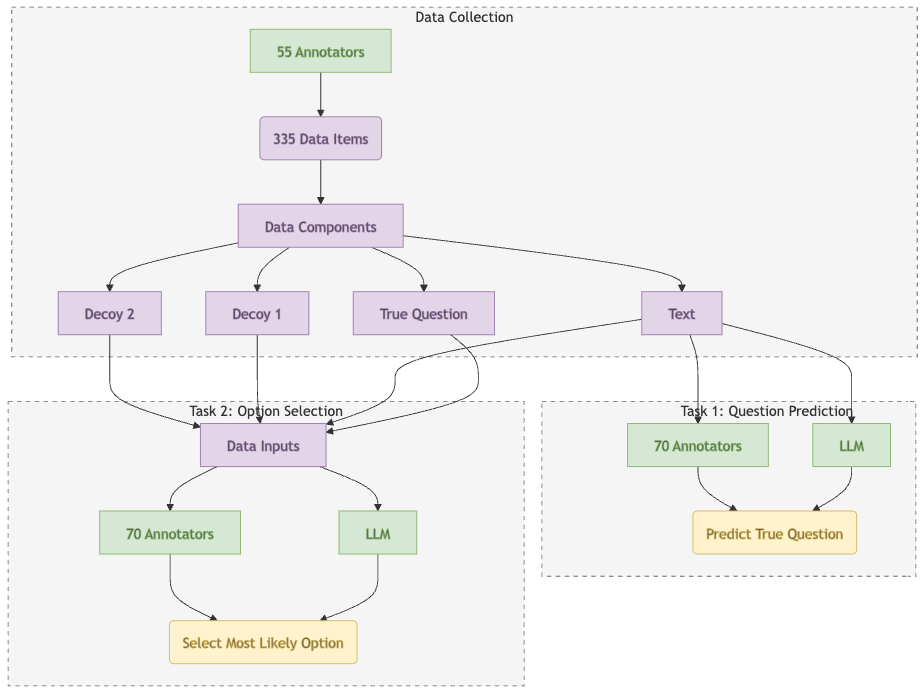}
    \caption{Experimental setup for evaluating LLM and human performance on question prediction. The experiment consists of two tasks: Task 1 (question prediction based on text alone) and Task 2 (selection of the most likely question from a set of options, given the text).}
    \label{fig:explanation}
\end{figure}

We propose a specific evaluation paradigm, which we term "\textbf{question archaeology}" (\autoref{fig:explanation}), that builds upon these ideas\footnote{Code and data will be publicly available.}. It evaluates the ability of LLMs to infer the underlying "genesis question" that motivated the writing of a given monologic text. This test acts like a reverse cloze test where we give the LLM a novel text (which was not published on the Internet before) and ask it to infer the single, authentic question that led the author to write the text. By analyzing how accurately LLMs can reconstruct this original question, we provide a novel benchmark for measuring the capacity for deep textual understanding and authorial intent recognition. This test challenges LLMs to move beyond surface-level comprehension and is, arguably, much harder than the classic question-and-answer test.

Our primary contributions are as follows:

\begin{itemize}
\item We propose a new evaluation paradigm, termed "question archaeology", which challenges LLMs to infer the original question that motivated the creation of a given text, rather than just factual recall.
\item We present a unique dataset of 335 newly commissioned texts, called "QuestionArchaeology", each paired with its ground-truth originating research question and two carefully crafted plausible distractor questions. The methodology employed in creating this dataset ensures its novelty and prevents contamination from pre-training data, thereby providing a robust benchmark for assessing genuine inferential abilities. We make the dataset publicly available.
\item This work presents the first evaluation of LLMs on the "question archaeology" task. 
\item We demonstrate advancements by evaluating the performance of older versus newer models on both a novel dataset and a previously unseen task, thereby showing genuine progress in their generalization capabilities beyond familiar scenarios. 
\item We provide a comparative analysis of Masked Language Models (MLMs) and autoregressive Large Language Models (LLMs) performance \textbf{against human annotators} on the same tasks. Our results reveal that current LLMs outperform humans in this specific domain.
\end{itemize}

\section{Related Work}

The task of inferring the motivation behind a text is related to several established research areas in NLP and discourse analysis. Our work builds on these foundations while carving out a specific, complementary niche focused on authorial intent.

The most direct parallel is \textbf{Question Generation (QG)}, where the goal is to generate a question given a passage. However, a key distinction lies in the objective. Standard QG aims to generate any plausible question that is answerable by the text \cite{du-etal-2017-learning}, and is often used for data augmentation for QA systems. In contrast, "question archaeology" is not about generating a diverse set of possible questions; it is a more constrained task of inferring the single, authentic, ground-truth question that actually motivated the text's creation. This shifts the focus from answerability to the recognition of authorial intent. Our task should also be distinguished from "reverse prompting", the task of inferring the prompt used to generate a specific text from a language model \cite{Morris2024LanguageModelInversion}. While both tasks reverse the typical input-output flow, question archaeology seeks to uncover the cognitive, human-centric motivation behind a human-authored text, whereas reverse prompting aims to deconstruct the machine-level instruction for a synthetic output.

Our work is also deeply grounded in theories of discourse structure that model communication as a series of questions and answers. Frameworks for identifying \textbf{implicit questions} \cite{fu-etal-2019-asking} and \textbf{Questions Under Discussion (QUD)} \cite{wu-etal-2023-qudeval} are highly relevant. QUD, in particular, offers a powerful formalism for modeling discourse coherence by treating each utterance as an attempt to answer a question on a structured stack. Similarly, frameworks for identifying \textbf{Dialogue Acts}, such as the Penn Discourse Treebank (PDTB) \cite{prasad-etal-2008-penn}, model the intention behind individual utterances within a broader dialogue.

While these frameworks are powerful for modeling the complex, turn-by-turn structure of discourse and utterance-level intent, especially in multi-party dialogues, our "question archaeology" task simplifies the problem to focus on a different, albeit related, goal. We justify our design choice by framing it as a complementary task focused specifically on uncovering the single, high-level \textbf{"genesis question"} for a complete, self-contained, monologic text. Rather than modeling the entire discourse stack or the function of each utterance, we aim to identify the overarching authorial purpose of the text as a whole. This provides a distinct, measurable, and challenging benchmark for evaluating an LLM's ability to reason about the holistic intent of an author, extending beyond the structural analysis offered by QUD and PDTB. Our work also connects to intent detection \cite{gupta-etal-2018-semantic-parsing} and argument mining \cite{lawrence-reed-2019-argument}, which similarly attempt to recover implicit structures from text, but not the genesis question.

\section{Data Collection}

There are no public datasets available for this kind of task. In addition, \textbf{existing public datasets would compromise} the integrity of the evaluation. Large language models trained on public data would have encountered both the texts and their motivating questions during pre-training, rendering it impossible to distinguish between genuine inference and simple retrieval of memorized associations. To address this challenge, we developed a novel dataset based on freshly commissioned texts, ensuring that the original research questions were explicitly recorded prior to the writing phase. Our dataset comprises 335 unique texts, with each text paired with its authenticated originating research question and two plausible but incorrect alternatives. By constructing the dataset in this manner, we can effectively measure a model's ability to discern the true question behind a text, rather than its capacity to match previously encountered patterns.

Our dataset was created with the help of 55 participants, each contributing an average of 6 samples. Prior to participation, all individuals provided informed consent, and were fully briefed on the project's objectives and procedures. All data was created inside the classroom by hand to ensure that no LLMs were used. The data collection process followed three steps:
\begin{itemize}
    \item First, participants formulated a research question of their choice (true question).
    \item Second, they wrote a text that addressed this question.
    \item Finally, they created two alternative questions that could plausibly, but incorrectly, be seen as the motivation behind their text (Other Question 1 and 2). To ensure quality, participants were asked to review their alternative questions to strike a balance - these questions needed to be related enough to the text to be credible, yet distinctly different from the original research question that motivated the writing.
\end{itemize}

\subsection{Model Evaluation}

Our research was structured around two distinct tasks. The \textbf{first task} evaluated the direct prediction of the motivating question from the text. The \textbf{second task} was a multiple-choice question requiring the selection of the correct motivating question from three options. To establish a robust human performance baseline for this second task and to calculate inter-annotator agreement, we collected 553 annotations from human raters. This number is greater than the 335 unique texts in our dataset because some texts were annotated by multiple raters. We made our code and dataset publicly available in a \href{https://github.com/AnonymousReview}{GitHub Repository}. 

We present here an example of an item in the dataset. The distractors were created by the original text authors to be deliberately plausible and challenging. Our aggregate analysis shows that, on average, the true question has a quantifiably higher semantic similarity to the text than the distractors (0.66 vs 0.57 and 0.54), indicating the task is well-defined, even if some examples are intentionally difficult. 

\begin{itemize}

\item The question that initiated the writing process (the annotator came up with the question himself):  Why do people get upset?

\item The text that was written based on that question: Sometimes, everything feels fine until something bad happen - for example, a harsh word, a missed promise, and so on. People carry much inside, including hopes, fears, frustrations. When something is not as expected, it can feel like the world is not listening. Anger or sadness can be a way of saying “This is not fair” or “I need someone to notice this/me”. Many times it is not about the event itself, but more about our feelings - about us being hurt, ignored, or misunderstood.

\item Two questions as decoys provided by the annotator: a) Why it is hard to stay calm when things go wrong? b) How do small misunderstandings escalate into big conflicts?
\end{itemize}

\begin{figure}[htbp]
    \centering
    \includegraphics[width=0.9\linewidth]{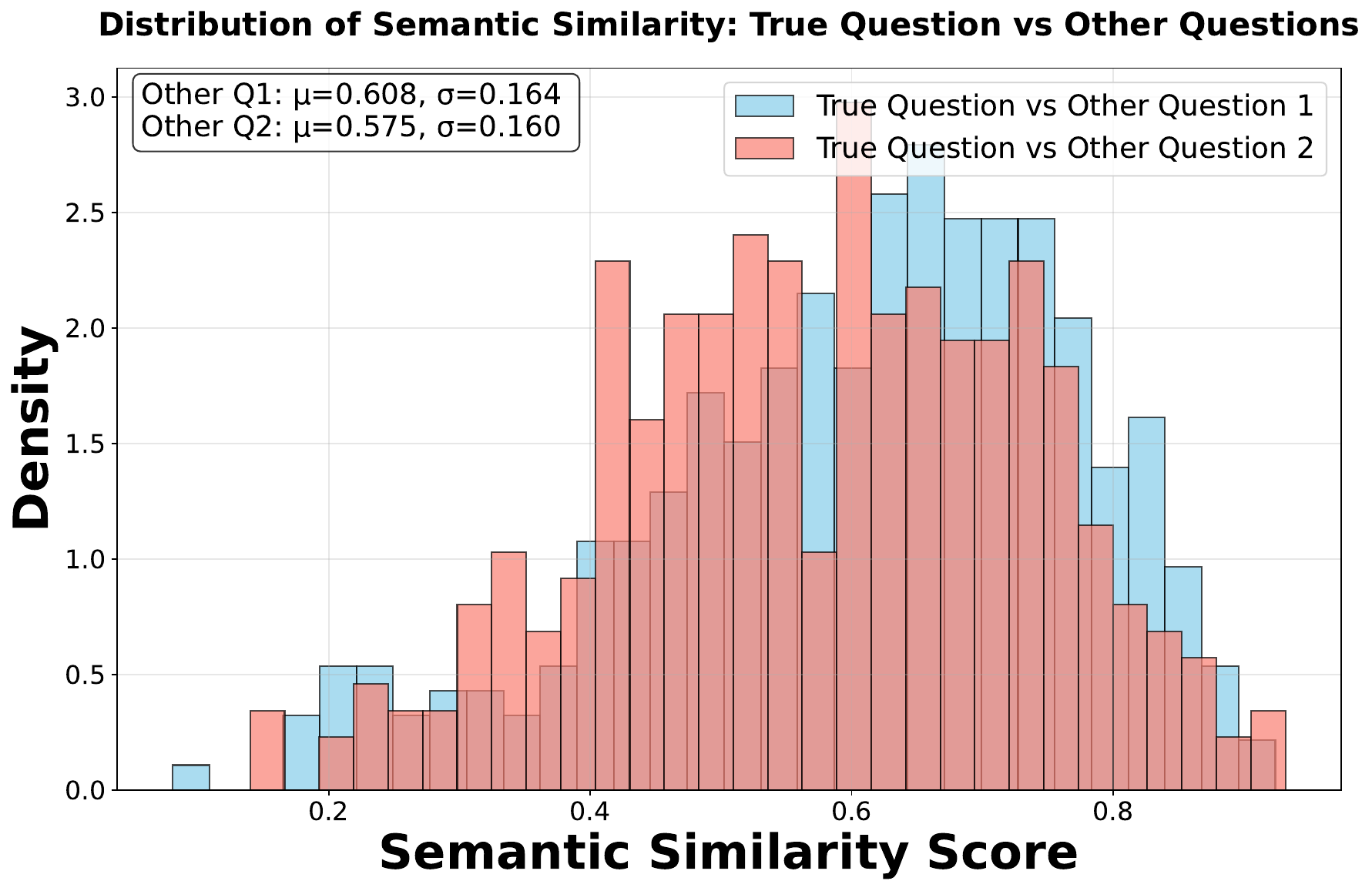}
    \caption{The semantic similarity scores between the true question and the two decoy questions, "Other Question 1" and "Other Question 2," show distributions that approximate a normal pattern. Other Question 1 exhibits a mean semantic similarity of 0.608 with a standard deviation of 0.164.
Other Question 2 has a slightly lower mean semantic similarity of 0.575, with a comparable standard deviation of 0.160.}
    \label{fig:question_2}
\end{figure}

Our participants represented a diverse group spanning multiple Mediterranean and Eastern European countries, including Italy, Romania, Greece, Morocco, Slovenia, and Egypt. The gender distribution showed a majority of female participants (63\%) compared to male participants (37\%), contributing to the diversity of perspectives in our dataset. All texts were written in English. Despite the annotators not being native English speakers, no language-related issues were observed. The analysis of inter-annotator agreement revealed an observed agreement proportion ($\bar{P}$) of 0.58. After accounting for an agreement expected by chance ($\bar{P}_e$) of 0.34, the Fleiss' Kappa coefficient ($K$) was 0.36. According to the benchmarks established by Landis and Koch \cite{landis1977measurement}, this Kappa value indicates fair to moderate agreement among the annotators. This moderate level of agreement is expected, as the interpretation of a text's underlying motivation is inherently subjective and tied to individual perspectives \cite{creanga-dinu-2024-designing}.

\begin{figure}[htbp]
    \centering
    \includegraphics[width=0.9\linewidth]{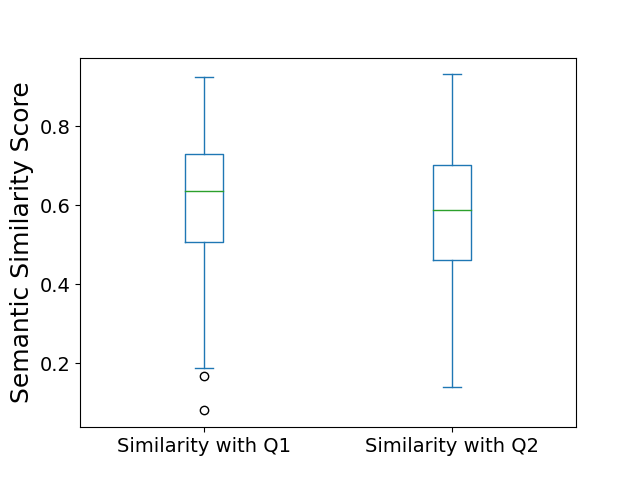}
    \caption{Distribution of Semantic Similarity between the true question and other questions. The data contains relatively few outliers. }
    \label{fig:boxplot}
\end{figure}

\begin{figure}[htbp]
    \centering
    \includegraphics[width=0.9\linewidth]{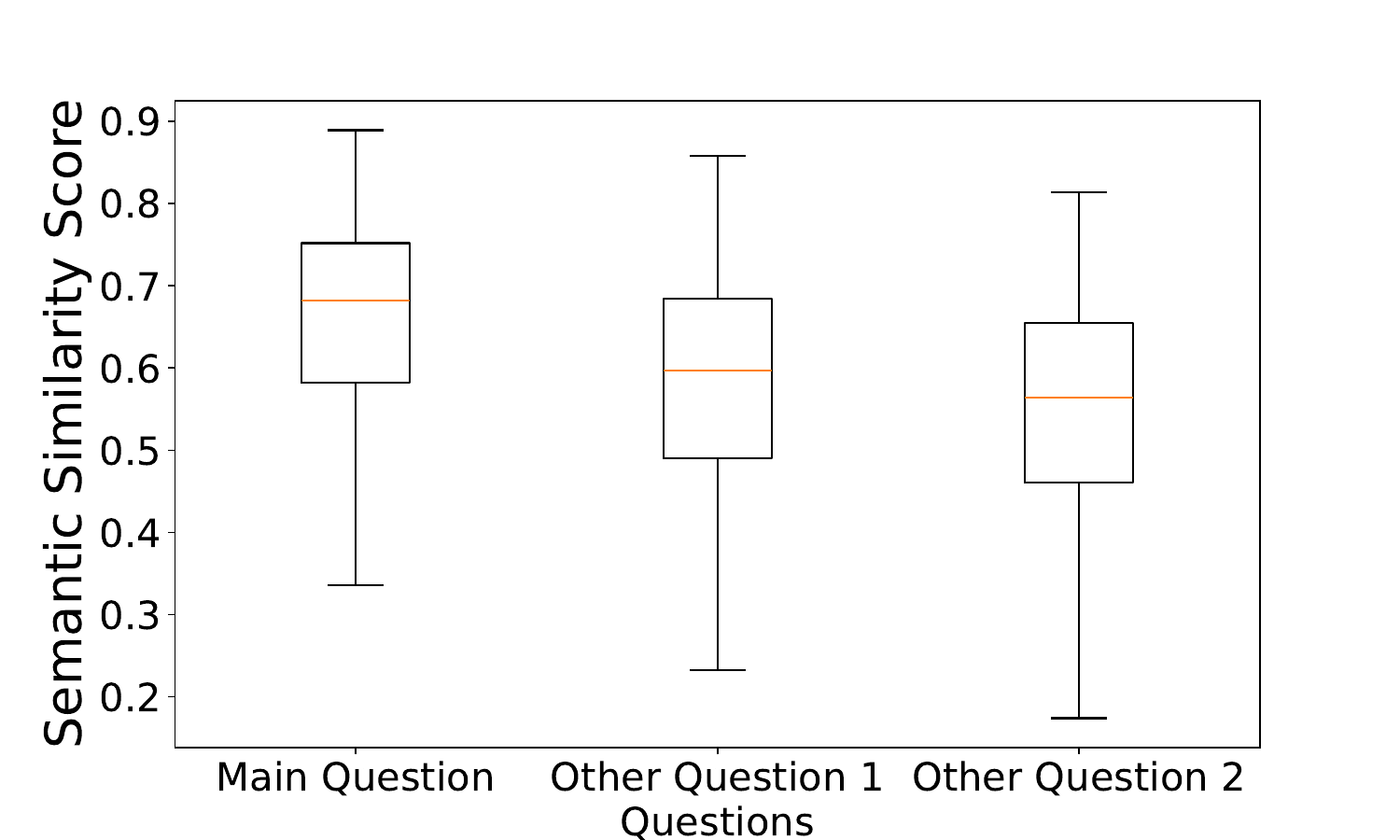}
    \caption{Distribution of Semantic Similarity between the text and each of the 3 questions. We can see a bigger similarity with the true question, but not by a lot: 0.66 vs 0.57 and 0.54.}
    \label{fig:choice_boxplot}
\end{figure}

Analysis of \textbf{semantic similarity} between the true questions and their alternatives reveals a pattern approximating a normal distribution. The mean similarity score is 0.57 for the second alternative question (\autoref{fig:question_2}) and 0.60 for the first alternative question. These moderate similarity scores indicate that participants successfully created alternative questions that maintain a balanced distance from the original - related enough to be plausible, yet sufficiently distinct to serve as meaningful distractors. The box-plot (\autoref{fig:boxplot}) reveals that the data contains relatively few outliers. The average semantic similarity between the text and the correct answer was 0.66, compared to 0.57 and 0.54 for the decoys (\autoref{fig:choice_boxplot}). The semantic similarity score was calculated using the Sentence Transformers model\cite{reimers-gurevych-2019-sentence}. 

Comparing how each alternative relates to the true question, we found a moderate positive correlation (Pearson r = 0.49) between their similarity scores. This indicates a tendency for both alternative questions to maintain comparable semantic distances from the original question. Question 1 appears to be semantically closer to the true question compared to Question 2, likely influenced by the order in which participants generated the alternatives. A paired t-test revealed a significant difference in the similarity scores between the two alternative questions (t = 3.66, p < 0.001), indicating that this difference in means (question 1 being closer to the true question than question 2) is unlikely to be due to random chance.

\subsection{Linguistic Analysis}

\begin{figure*}[htbp]
    \centering
    \includegraphics[width=0.9\linewidth]{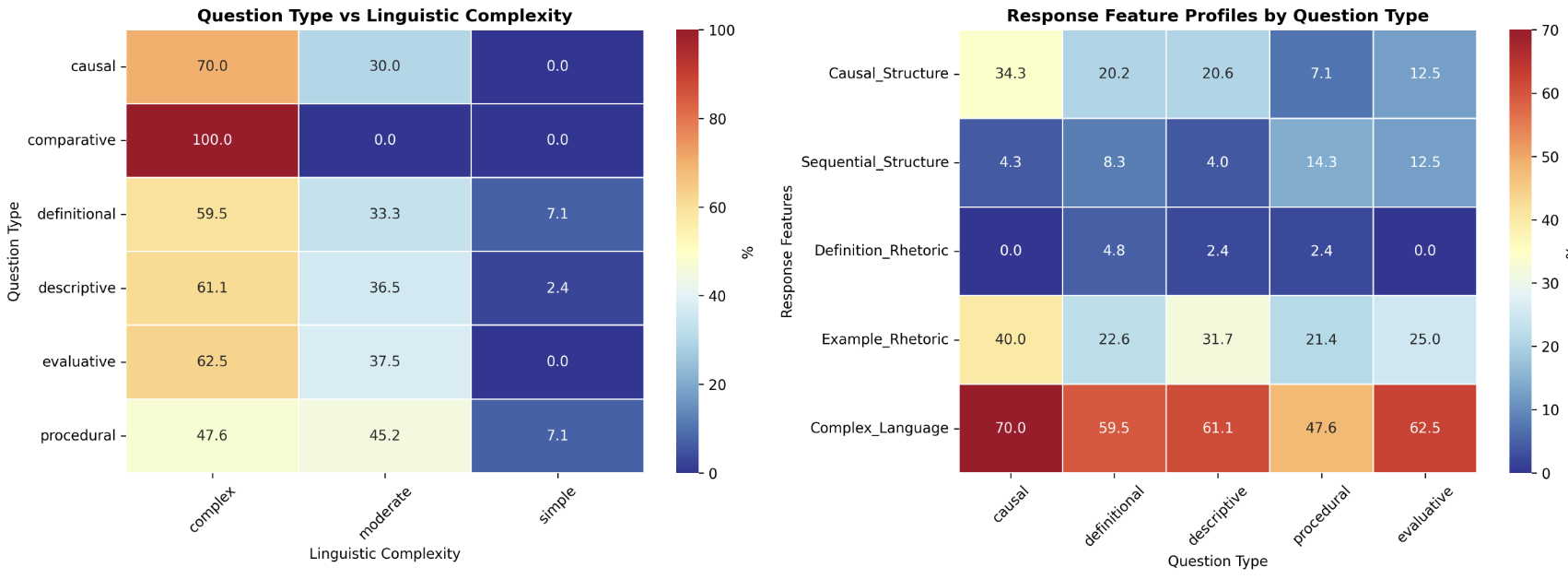}
    \caption{Heatmap showing that question type serves as a strong predictor of response structure, with different interrogative forms activating distinct cognitive and linguistic patterns in human responses.}
    \label{fig:heatmap}
\end{figure*}

The linguistic analysis in \autoref{fig:heatmap} reveals that question types fundamentally shape discourse organization in predictable ways, with causal questions ("Why") generating the most linguistically complex responses. These texts contain significantly more causal conjunctions (2.21 expressions per text compared to 1.00-2.06 for other types) and employ sophisticated argumentative structures moving from problem statements through causal explanations to consequences. Meanwhile, definitional questions produce categorical discourse with clear topic-comment structures and explanatory markers like "for example," while procedural questions generate sequential patterns dominated by temporal markers ("first," "then," "next") rather than causal reasoning.

The syntactic signatures align closely with discourse functions, creating distinct textual fingerprints for each question type. Causal texts favor complex sentences with subordinate clauses rich in markers like "because," "due to," and "therefore," while definitional texts rely heavily on copulative constructions (X is Y) and categorical statements. Procedural texts show more imperative and conditional structures, reflecting their instructional nature. Despite similar sentence counts across types, causal questions generate slightly longer texts (81.9 words versus 75-77 words), suggesting that explaining causation requires more elaborate linguistic resources than other cognitive tasks like definition or description.

\begin{figure}[htbp]
    \centering
    \includegraphics[width=0.9\linewidth]{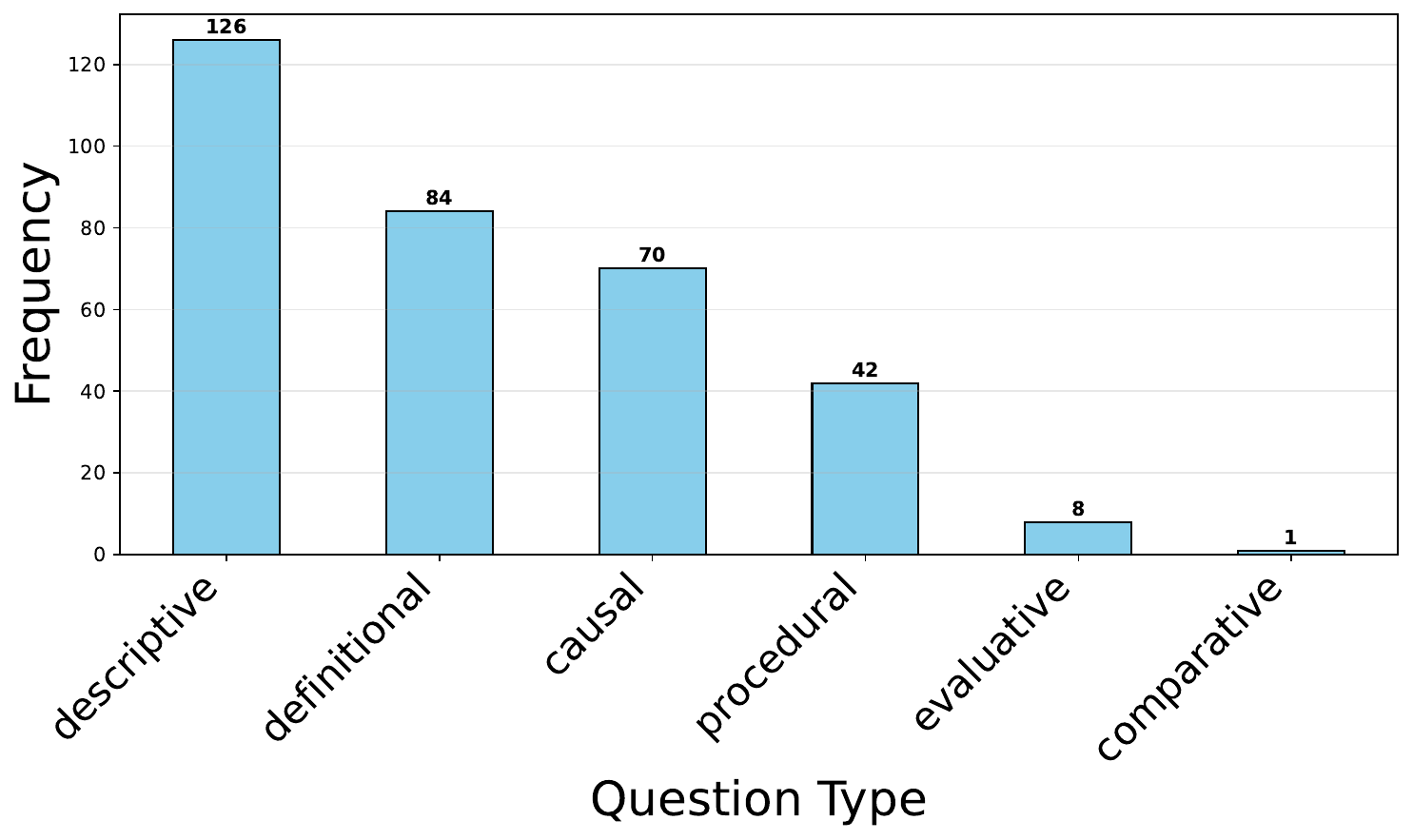}
    \caption{Distribution of question types by counts. Most questions are descriptive.}
    \label{fig:question_distribution}
\end{figure}

\begin{figure}[htbp]
    \centering
    \includegraphics[width=0.9\linewidth]{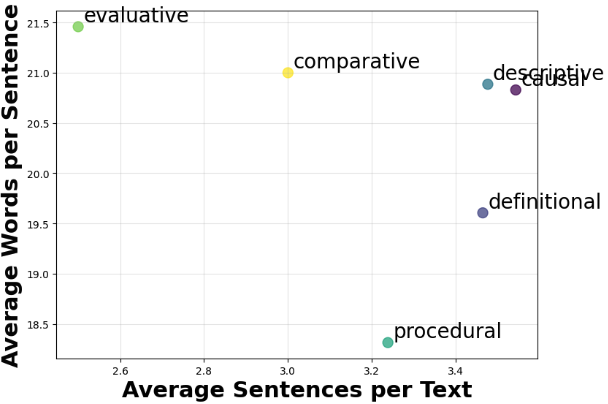}
    \caption{Distribution of question types by counts. Most questions are descriptive.}
    \label{fig:words_sentence}
\end{figure}

Quantitative analysis \autoref{fig:question_distribution} and \autoref{fig:words_sentence} shows causal questions demonstrate high usage of causal discourse structures (34.3\%), definitional questions favor narrative rhetoric and simple structure, and procedural questions consistently employ sequential organization patterns. \autoref{fig:heatmap} indicate that question type serves as a strong predictor of response structure, with different interrogative forms activating distinct cognitive and linguistic patterns in human responses. Question type appears to directly shape how knowledge is expressed, with each type triggering predictable discourse strategies that reflect the underlying cognitive processes required to construct responses.

\section{Results}

To establish a baseline, we implemented an XGBoost classifier using TF-IDF features (for task 2). We experimented with various preprocessing and feature engineering steps, such as limiting TF-IDF features (max\_features=5000), incorporating n-grams (ngram\_range=(1,2) and ngram\_range=(1,3)), applying lemmatization, and filtering high-frequency terms (max\_df=0.95 to ignore terms appearing in > 95\% of documents). The optimal configuration utilized TF-IDF (max\_features=5000) with bi-grams (ngram\_range=(1,2)), yielding an average F1-score of 0.35, which is only slightly above the threshold for random chance.

\subsection{Assessing LLM Performance}

Our experiment aimed to evaluate for the first time the ability of LLMs to identify the original question that prompted a given piece of text and to measure the progress between different versions. We compared four versions of the Gemini models \cite{geminiteam2023gemini}: Gemini Flash 2.0, Gemini Flash 2.0 Thinking (which incorporates enhanced reasoning capabilities), Gemini Flash 2.5 and Gemini Pro 2.5. This comparison allowed us to observe the evolution of these models in finding the true questions.

\begin{table}[h]
\caption{Comparison of LLMs and human performance on task 1, based on F1 Macro scores.}
\begin{tabular*}{\hsize}{@{\extracolsep{\fill}}ll@{}}
\toprule
Model  & F1 score \\
\midrule
\textbf{Human raters} & \textbf{0.66} \\
Gemini Flash 2.0  & 0.68 \\
Gemini Flash 2.0 Thinking & 0.71 \\
Gemini Flash 2.5 & 0.72 \\
\textbf{Gemini Pro 2.5} & \textbf{0.75} \\
\bottomrule
\label{table:llm1}
\end{tabular*}
\end{table}

To assess this, we designed a two-pronged evaluation approach. In the first task, we challenged the models to directly predict the original question based solely on the text provided by the authors. The second task involved a multiple-choice format, where the models were presented with the generated text and a list of three questions, one of which was the original question, and asked to select the correct one.

\begin{figure}[htbp]
    \centering
    \includegraphics[width=0.9\linewidth]{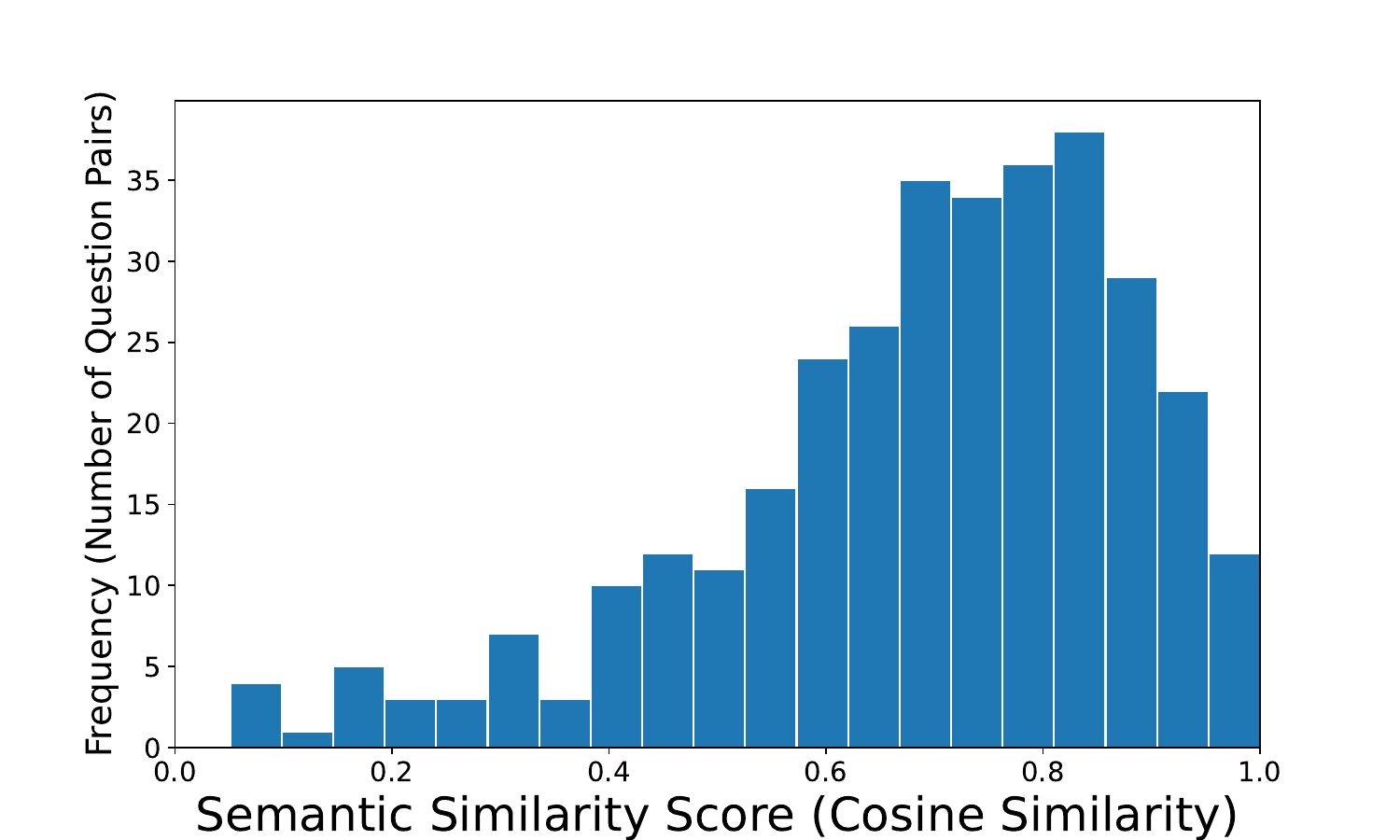}
    \caption{Distribution of Semantic Similarity between the true question and the question predicted by Gemini Flash 2.0.}
    \label{fig:early_question_gemini}
\end{figure}

\begin{figure}[htbp]
    \centering
    \includegraphics[width=0.9\linewidth]{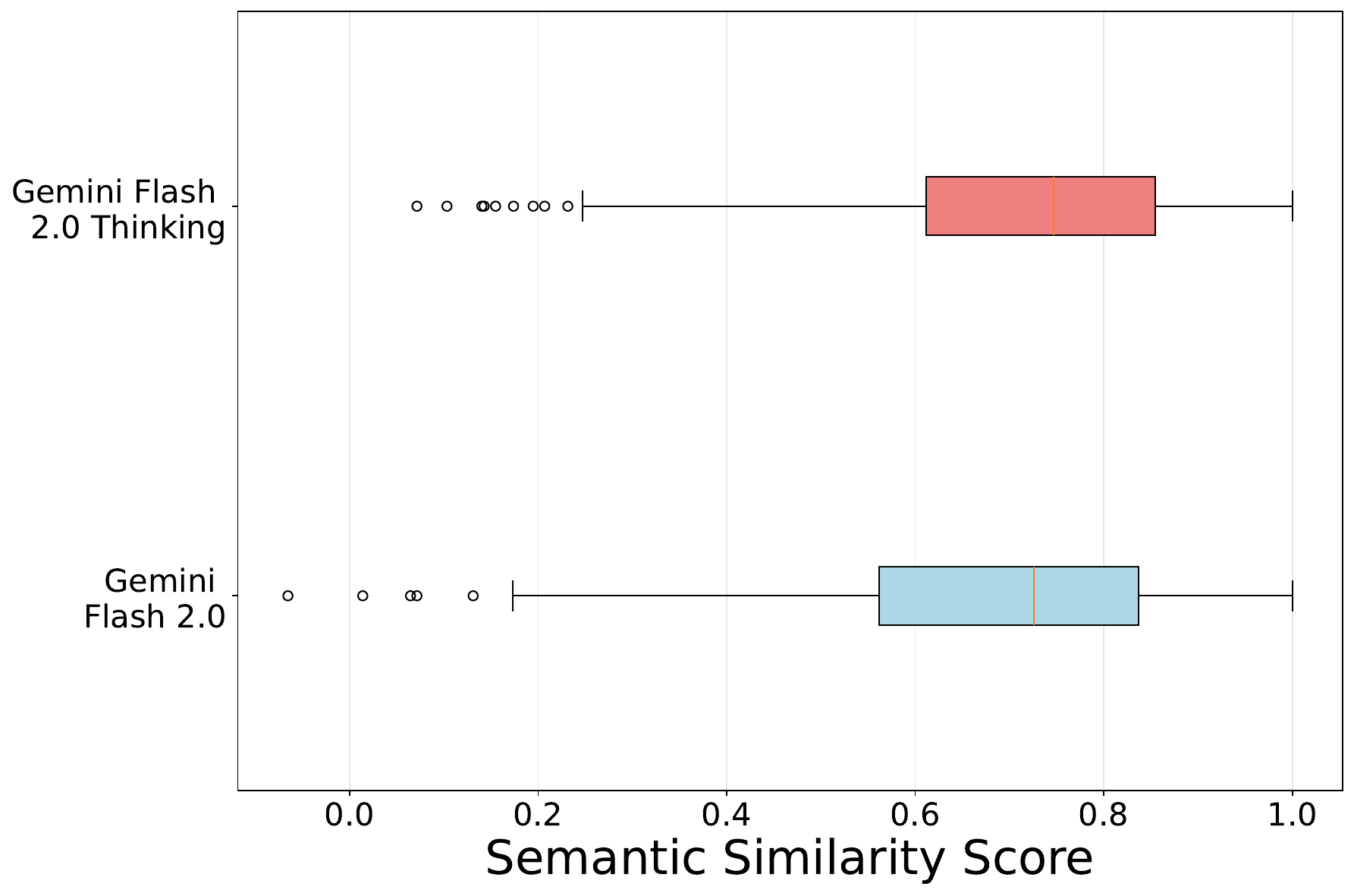}
    \caption{Plot showing semantic similarity scores for Gemini Flash 2.0 and  Gemini Flash 2.0 Thinking. For Gemini Flash Thinking 2.0 the percentage of question pairs with similarity > 0.8 is 38.07\%, and for Gemini Flash 2.0 the percentage of question pairs with similarity > 0.8: 33\%.}
    \label{fig:question_gemini_box}
\end{figure}

Focusing on the first task of direct question prediction, the earlier version of the model, Gemini Flash 2.0, demonstrated a \textbf{strong baseline performance}, as seen in \autoref{table:llm1}. It achieved an average semantic similarity score of approximately 0.68 (\autoref{fig:early_question_gemini}). This score suggests that, on average, the questions generated by the model exhibited a considerable degree of semantic overlap with the actual questions posed to the annotators. Further analysis revealed that in 12 instances, the model precisely predicted the original question (achieving a perfect similarity score of 1). Moreover, in a significant portion of cases (33\%), the semantic similarity between the predicted and actual questions exceeded 0.8 (\autoref{fig:question_gemini_box}), indicating a high level of accuracy in capturing the essence of the original inquiry.

\begin{figure}[htbp]
    \centering
    \includegraphics[width=0.9\linewidth]{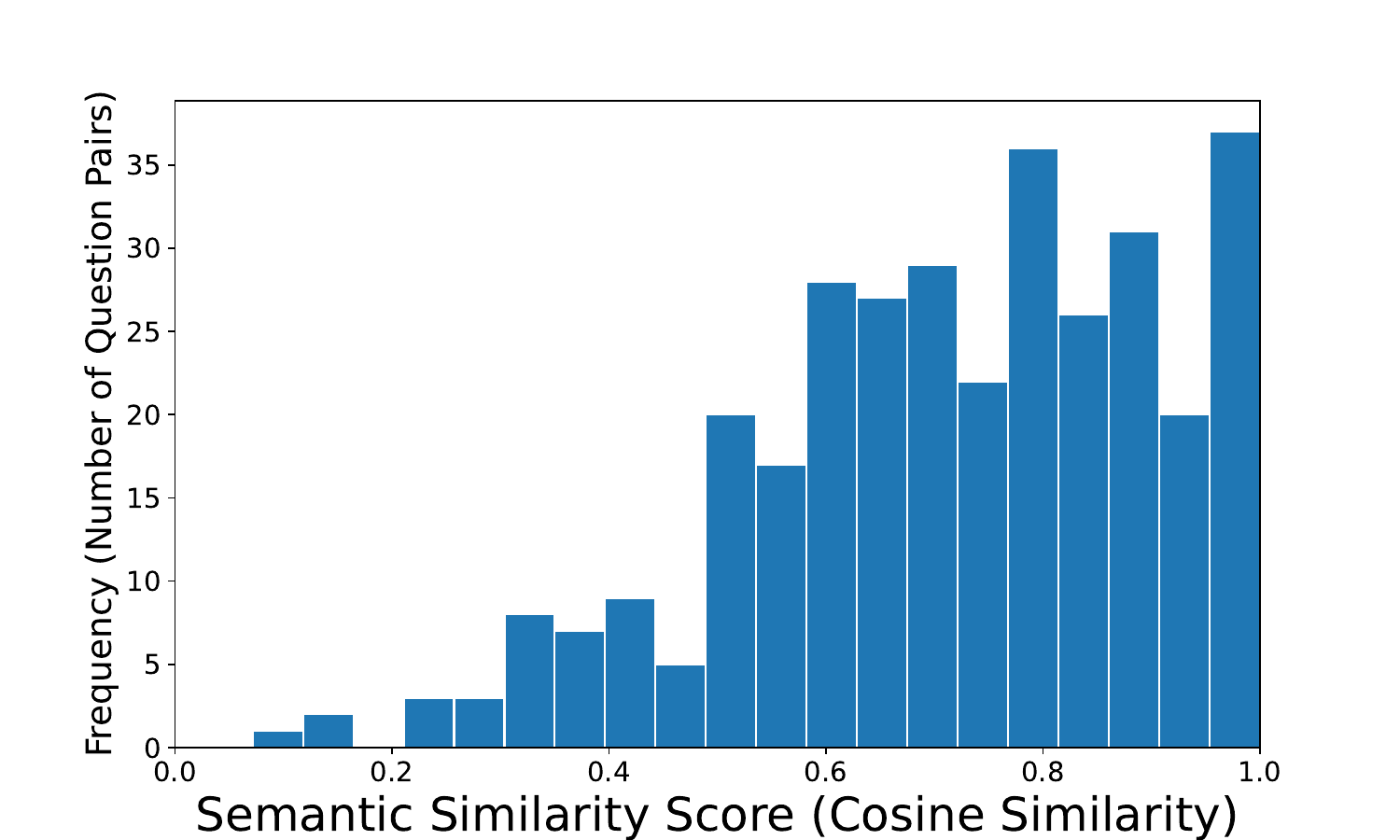}
    \caption{Distribution of Semantic Similarity between the true question and the question predicted by Gemini Flash 2.0 Thinking. The model  demonstrated an increased ability to precisely identify the original question, achieving a perfect similarity score of 1 in 30 cases. }
    \label{fig:question_gemini}
\end{figure}

The thinking version of the model, Gemini Flash 2.0 Thinking, showed a notable improvement. It achieved an even \textbf{higher average semantic similarity score} of approximately 0.71 (\autoref{fig:question_gemini}). Interestingly, the median similarity score (0.73) was slightly higher than the mean, indicating that the distribution of similarity scores was positively skewed, with a greater number of texts leading to higher similarity predictions. The model also demonstrated an increased ability to precisely identify the original question, achieving a \textbf{perfect similarity score} of 1 in 30 cases. Furthermore, the percentage of cases where the semantic similarity was greater than 0.8 rose to 38\% (\autoref{fig:question_gemini_box}), highlighting a consistent enhancement in the model's ability to understand the motivations of the authors and reproduce the underlying questions. We also explored the confidence levels reported by the Gemini Flash 2.0 Thinking model for its question predictions. While the model generally expressed confidence in its answers, the correlation between its confidence score and the actual semantic similarity achieved was only low to medium (a Pearson correlation coefficient of 0.25). This suggests that while the model provides a confidence estimate, this estimate does not strongly align with the objective measure of how semantically similar its prediction is to the true question. This observation warrants further investigation to understand the factors influencing the model's confidence calibration and whether improvements can be made to better reflect the accuracy of its predictions. Both Gemini 2.5 models performed strongly, with the Pro version achieving the highest F1 score (0.75), while the Flash version achieved 0.73. Notably, the Gemini 2.5 Pro score was almost 10 points above human raters. In assessing semantic similarity greater than 0.8, the Pro model obtained an impressive 46\% of cases, compared to 37\% for the 2.5 Flash model. 

We explored alternative strategies to improve our score, but these efforts were unsuccessful. One such approach involved first summarizing the text and then having the model predict the question based on the summary. This method yielded results approximately 4\% worse than our standard approach, with scores of 0.69 for Gemini Flash 2.5 and 0.71 for Gemini Pro 2.5.

This initial analysis of the first task clearly demonstrates an evolution in the ability of the models in the Gemini series to identify the original question from a given text.

The second task required models to choose from three options (\autoref{table:llm2}). Initial results showed strong performance, with Gemini Flash 2.0 achieving 0.649 accuracy and Gemini Flash 2.0 Thinking reaching 0.653, both surpassing other MLMs and human raters. The Gemini 2.5 series demonstrated substantial improvement, yielding F1 scores of 0.734 (Gemini Flash 2.5) and 0.764 (Gemini Pro 2.5). Lower F1 scores were obtained by the open-source models Mistral \cite{jiang2023mistral} (0.57) and Qwen \cite{Bai2023} (0.59). Open source models were also used in a zero-shot setting with no fine-tuning. Interestingly, we found no correlation between the semantic similarity of the options and the models' selections.

\begin{table}[h]
\caption{Comparison of LLMs and human performance on task 2, based on F1 Macro scores.}
\begin{tabular*}{\hsize}{@{\extracolsep{\fill}}ll@{}}
\toprule
Model  & F1 score \\
\midrule
XGBoost & 0.35 \\
Mistral:7B & 0.57 \\
Qwen 2.5:14B & 0.59 \\
\textbf{Human raters} & \textbf{0.60} \\
Gemini Flash 2.0  & 0.65 \\
Gemini Flash 2.0 Thinking & 0.65 \\
Gemini Flash 2.5 with summarization & 0.69 \\
Gemini Pro 2.5 with summarization & 0.71 \\
Gemini Flash 2.5 & 0.73 \\
\textbf{Gemini Pro 2.5} & \textbf{0.76} \\
\bottomrule
\label{table:llm2}
\end{tabular*}
\end{table}

We minimized variability by setting the LLM temperature to 0 and running each test three times. The second task demonstrated high consistency, with only four cases requiring a majority vote to determine the final output. We didn't fine-tune and all LLM experiments were done in a zero-shot setting.

\subsection{Assessing MLM Performance}
While LLMs currently dominate many NLP tasks, we investigated the competitiveness of Masked Language Models (MLMs) on the multiple-choice task. We experimented with several prominent MLMs: DeBERTa V3 \cite{He2020}, BERT large \cite{Devlin2018}, XLM-RoBERTa \cite{Conneau2019}, and the recent large-scale ModernBERT \cite{Warner2024}. We used them only for task 2, the classification one, and on top of these models we added a classification head. This head consisted of a normalization layer for input stabilization, a Tanh activation function for non-linearity, a dropout layer for regularization, and a final layer mapping outputs to one of three classes. Our fine-tuning employed a two-stage strategy: initially, we froze all transformer layers and trained only the classification head for 2 epochs with a learning rate of 3e-4; subsequently, we unfroze the final transformer layer and continued training both the head and this layer for 1 epoch at a reduced learning rate of 2e-4. As the results in \autoref{table:masked} indicate, MLMs performed significantly worse than LLMs. This presents an interesting contrast to other NLP tasks, such as sentence-level emotion recognition, where fine-tuned MLMs have been shown to be highly competitive against zero-shot autoregressive models \cite{creanga-dinu-2024-transformer}. This disparity likely stems from fundamental differences in their pre-training: MLMs focus on local context via masked word prediction, which may be less effective for grasping global text purpose or authorial intent compared to the next-token prediction objectives common in LLM training. Furthermore, simply fine-tuning on this small dataset (we used a 40-60 split) an MLM with a classification head might not instill the nuanced understanding required to differentiate the true motivating question from plausible distractors, a task potentially better suited to the scale and inherent capabilities of LLMs. To enhance performance, we used an ensemble of 3 models, ModernBERT, XLM-RoBERTa, and DeBERTa, with a soft voting strategy (averaging class probabilities). The resulting F1-score for the ensemble was 0.52, representing only a slight improvement over the baseline performance of ModernBERT.

\begin{table}[h]
\caption{Comparison of MLMs performance on task 2, based on Precision, Recall, and F1 Macro scores.}
\begin{tabular*}{\hsize}{@{\extracolsep{\fill}}llll@{}}
\toprule
Model  & Precision & Recall & F1 score \\
\midrule
BERT large & 0.51 & 0.45 & 0.47 \\
DeBERTa V3  & 0.50 & 0.48 & 0.48 \\
XLM-RoBERTa  & 0.53 & 0.50 & 0.50 \\
ModernBERT & 0.52 & 0.53 & 0.51 \\
Ensemble & 0.52 & 0.54 & 0.52 \\
\bottomrule
\label{table:masked}
\end{tabular*}
\end{table}

\subsection{Assessing Human Performance}

In the first task, we asked human annotators to infer the original question solely from the provided text. On average, the semantic similarity between their inferred questions and the actual questions was approximately 0.66, with a median of 0.68 (\autoref{table:llm1}). This indicates a \textbf{moderate} level of semantic overlap but falls short of the performance achieved by the LLM. Furthermore, only 21.61\% of the inferred question pairs exhibited a high degree of similarity (score > 0.8). Notably, in just 20 instances did the annotators correctly guess the exact question.

For the second task, where annotators were presented with three potential questions and asked to select the one that best corresponded to the text, the human accuracy rate was 0.60 (\autoref{table:llm2}). This is a big difference compared to the LLM's accuracy of 0.76. Examining the distribution of choices, we observe a relatively even spread across the three options. While question 2 received slightly fewer selections, the difference was not substantial, suggesting that annotators weren't drawn to a particular order.

\section{Discussion}

The results of our evaluation suggests that LLMs (the Gemini models tested) are better than humans at inferring the questions behind a particular text. In the direct question prediction task, the progression from the earlier Gemini Flash 2.0 model to the more recent versions is clearly demonstrated by the \textbf{significant improvement} in average semantic similarity and the substantial increase in the number of exact question matches. This indicates that the enhanced reasoning capabilities incorporated into the newer model have indeed improved its ability to understand the underlying intent that prompted the given text. The 4 models were released a couple of months apart, but the difference is significant.

Interestingly, there was a bigger difference between the best model and human raters in task 2 (16\%), than in task 1 (9\%). This might indicate that while humans are reasonably adept at formulating a potential question, they are more susceptible than LLMs to being misled by carefully crafted distractors in a multiple-choice setting. The low to medium correlation observed between the model's confidence scores and the actual semantic similarity achieved in the direct prediction task raises important questions about the calibration of confidence in LLMs. They appear to be too confident in their predictions. While the model expresses a degree of certainty in its predictions, this confidence was always bigger than 80\% and \textbf{does not align with accuracy}, indicating a potential area for future research and improvement.

Still, the results suggest that, at least for this specific task, current state-of-the-art LLMs demonstrate a superior ability to understand the nuances of text and identify the underlying questions that motivated its creation. This highlights the potential of LLMs as valuable tools for tasks involving information retrieval and understanding the relationship between questions and answers.

\section{Conclusion and Future Work}

In this paper, we introduced "question archaeology", a novel task designed to evaluate the ability of human and LLMs to infer the underlying questions that motivated the writing of a given text. This task challenges LLMs to move beyond text comprehension and delve into the author's intent, arguably representing a more profound level of understanding than traditional question-answering. Our evaluation of four versions of the Gemini models demonstrated a clear progression in their ability to perform this task. The newer versions exhibited significant improvements in directly predicting the original question, achieving higher semantic similarity scores and a greater number of exact matches compared to its predecessors. Interestingly, we observed a notable discrepancy between the high confidence expressed by the newer models and the actual semantic similarity of its predictions, highlighting a potential area for improvement in confidence calibration. Furthermore, our comparison with human annotators revealed that the evaluated LLMs currently \textbf{outperform humans} in both direct question inference and question selection.

Despite promising results, future work should address the poor confidence calibration in LLMs, perhaps through uncertainty estimation techniques. A broader evaluation across diverse model architectures and sizes is also needed to identify how textual characteristics—such as length, domain, and question complexity—influence the difficulty of the "question archaeology" task and reveal the limits of model reasoning.

Beyond semantic similarity, future work could explore the use of more nuanced \textbf{evaluation metrics} that capture different aspects of the relationship between the predicted and original questions, such as topical relevance or the level of detail. Finally, investigating the practical applications of "question archaeology" in areas like information retrieval, where understanding the user's underlying intent is fundamental, or in educational settings to assess comprehension, could further highlight the significance of this research direction. Expanding our novel dataset with a larger and more diverse collection of texts and questions.



\section*{Limitations}

Our dataset is fairly small, comprising 335 newly commissioned texts and, on top of those, 553 annotated samples for human evaluation. While the novelty of our dataset ensures no contamination from pre-training data, its limited size might restrict the generalizability of our findings to a broader range of text types, domains, and question complexities. The participant pool, while diverse geographically, had a gender imbalance (63\% female), which could potentially introduce a bias in the types of questions formulated and the writing styles employed. Furthermore, all texts were written in English, limiting the applicability of our benchmark to other languages. While we took measures to ensure the quality of the distractor questions, their plausibility is inherently subjective and could influence the difficulty of the multiple-choice task. Finally, our evaluation primarily focused on one family of LLMs (Gemini), 2 open source ones (Mistral and Qwen), and 4 BERT-based models, so further research is needed to assess the performance of other architectures and training paradigms on the question archaeology task.

\section{Ethics Statement}
The manual labeling was carried out by volunteering students. Participation was optional, and alternative activities of comparable effort were available for those who chose not to annotate. All participants provided written informed consent prior to annotation, confirming their understanding of the project's purpose, the voluntary nature of their participation, their right to withdraw at any time without penalty. We obtained all necessary ethical approvals for this study. The collected annotations will be released publicly as part of the dataset, ensuring contributor anonymity by removing any links between individual annotators and specific labels. We will release our data and code under the CC BY-NC-SA 4.0 license.

\section{Acknowledgements}

This research is supported by a grant from Accenture Lab.






\bibliographystyle{iccc}
\bibliography{iccc}

\end{document}